\documentclass{article}

\usepackage[preprint, nonatbib]{neurips_2022} 

\usepackage[utf8]{inputenc} 
\usepackage[T1]{fontenc}    
\usepackage{hyperref}       
\usepackage{url}            
\usepackage{booktabs}       
\usepackage{amsfonts}       
\usepackage{nicefrac}       
\usepackage{microtype}      
\usepackage{xcolor}         

\usepackage{caption}
\usepackage{subcaption} 
\usepackage{graphicx}
\graphicspath{{./figs/}}
\usepackage{float}

\newtheorem{conjecture}{Conjecture}

\title{Learning Robust Representations Of Generative Models Using Set-Based Artificial Fingerprints} %

\author{
Hae Jin Song$^{1,2}$ \qquad Wael AbdAlmageed$^{1,2,3}$\\
$^1$ USC Information Sciences Institute \\
$^2$ Visual Intelligence and Multimedia Analytics Laboratory \\
$^3$ USC Ming Hsieh Department of Electrical and Computer Engineering \\
\tt\small  \{haejinso, wamageed\}@isi.edu}

\begin{document}

\maketitle

\begin{abstract}
With recent progress in deep generative models, 
the problem of identifying synthetic data and 
comparing their underlying generative processes 
has become an imperative task for various reasons, including fighting visual misinformation and source attribution.
Existing methods often approximate the distance between the models via their sample distributions. 
In this paper, we approach the problem of fingerprinting generative models by learning representations that encode the residual artifacts left by the generative models as unique signals that identify the source models.  We consider these unique traces (a.k.a. ``artificial fingerprints") as representations of generative models, and demonstrate their usefulness in both 
the discriminative task of source attribution and 
the unsupervised task of defining a similarity between the underlying models. 
We first extend the existing studies on fingerprints of GANs to four representative classes of generative models (VAEs, Flows, GANs and score-based models), and demonstrate their existence and attributability.
We then improve the stability and attributability of the fingerprints by proposing a new learning method based on set-encoding and contrastive training.
Our set-encoder, unlike existing methods that operate on individual images, learns fingerprints from a \textit{set} of images. 
We demonstrate improvements in the stability and attributability through comparisons to state-of-the-art fingerprint methods and ablation studies.
Further, our method employs contrastive training to learn an implicit similarity between models.  
We discover latent families of generative models using this metric in a standard hierarchical clustering algorithm. 
\end{abstract}

\section{Introduction}
Recent advances in deep generative models have enabled generation of high-quality samples that are often indistinguishable from  real data by human perception. 
Particularly in computer vision, we have seen astonishing qualities of images being synthesized in high-resolutions, over diverse domains including faces, natural scenes and artistic creations \cite{Park2019SemanticIS, Karras2020AnalyzingAI, Karras2021AliasFreeGA},
as well as videos that reproduce temporal and semantic consistency and realism  \cite{Suwajanakorn2017SynthesizingO, Korshunov2018DeepFakesAN}. 
This success  has been largely enabled by Generative Adversarial Networks (GANs) \cite{Goodfellow2014GenerativeAN}
and its subsequent variants, such as DCGAN \cite{Radford2016UnsupervisedRL}, Wasserstein GAN \cite{Arjovsky2017WassersteinG, Gulrajani2017ImprovedTO, Petzka2018OnTR}, MMDGAN \cite{Binkowski2018DemystifyingMG}, and StyleGAN \cite{Karras2019ASG, Karras2020AnalyzingAI}.
More recently, however, models from other classes of generative models such as NVAE \cite{Vahdat2020NVAEAD}, GLOW \cite{Kingma2018GlowGF} and LSGM \cite{Song2021ScoreBasedGM} have started to achieve high-quality photorealism comparable to that of GANs, diversifying the set of generative models capable of realistic image synthesis.

Despite the wide variety of models being proposed, there is still no method that can characterize how and in what ways these models are different. 
Our work addresses the problem of identifying and comparing generative models based on their samples.
In particular, we approach this problem from the perspective of  \emph{artifical fingerprints}, which refers to the unique traces of computations left on the images by a given generative model.
We focus on two particular challenges brought about from the advancement and diversification of generative models, where the fingerprinting approach is particularly relevant.

The first challenge  is the difficulty of attributing synthetic media to their source model.  
As generative models become more accessible and easier to train and modify, 
the malicious use of this technology has also become prevalent in, for instance,  
creation of fake news \cite{Zhou2018FakeNA}, 
blackmailing \cite{Mirsky2021TheCA}, 
misrepresention of political figures \cite{Chesney2018DeepFA}, 
and digital copyright infringement \cite{Floridi2018ArtificialID, Pavis21RebalacingOR}.
Such misuses increase the risk of visual misinformation and threaten public safety and trust \cite{Chesney2018DeepFA,Nguyen2020DeepLF}. 
It is thus imperative to develop an automated system that can effectively detect synthetic visual media and attribute their sources.
To address this problem, a line of work in \emph{fingerprint}ing GANs has been recently evolved, in which GAN-generated images are the results of a large number of fixed filtering and non-linear processes, which result to common and stable patterns on the images they generate. 
Based on this idea, \cite{Marra2019DoGL, Yu2019AttributingFI, Asnani2021ReverseEO} use either hand-crafted or learned fingerprints to perform attribution for multiple GAN architectures and demonstrate high classification accuracies. 
However, their studies are limited to GANs, and as more generative models with comparable photorealism have been developed, there is a need to update the studies to answer whether such fingerprints exist for a larger class of generative models.
Our work therefore extends the hypothesis on GAN fingerprints to a generalized hypothesis on artificial fingerprints and studies the fingerprints of four representative classes of generative models: VAEs, Flow models, GANs, and score-based models. 
We introduce a new dataset of generative models that includes various state-of-the-art models from all four classes, which forms the basis of our analysis on artificial fingerprints. 
This dataset contains four types of VAEs, 14 types of GANs, two flow-based models and three types of score-based models which are all trained on the same training data (CelebA \cite{liu2015faceattributes}) with unified data-processing steps. 
This consistency allows for the comparison of fingerprints to be attributed to the differences in the models, rather than a combination of models and training datasets, as done in previous works \cite{Yu2019AttributingFI, Marra2019DoGL, Asnani2021ReverseEO}. 

Another challenge that has risen from the diversification of generative models is the difficulty of quantifying their similarities based only on their samples. 
Visual examination of the samples is expensive and limited, especially when the quality of samples by different models is comparable. 
As an alternative, many evaluation methods have been proposed over the last decade to automate the quantitative comparisons.
Most popular methods \cite{Heusel2017GANsTB, Binkowski2018DemystifyingMG}, share in common that they approximate the distance between the models (i.e., their sampling distributions) using semantic features of the samples.
For instance, Fréchet Inception Distance (FID) \cite{Heusel2017GANsTB} uses the Inception features from Inception v3 \cite{Szegedy2015GoingDW} with an additional assumption that the distributions of the features are unimodal Gaussian.
Kernel Inception Distance (KID) \cite{Binkowski2018DemystifyingMG} also uses the Inception features, but approximates the divergence using the squared maximum mean discrepancy (MMD) with a polynomial kernel.
However, since most state-of-the-art models achieve high-quality images that are comparably photorealistic, comparing these models based on their \textit{semantic} features may not be feasible. 
Instead, we approach this problem by turning to the \textit{non-semantic} residual artifacts left by generative models as unique signatures that reflect the similarities between the source models. 
To learn the similarity between models implicitly through the fingerprint representations, 
we employ contrastive training,
which encourages the embeddings of images from the same generative model to be close in the embedding space,
and those from different models to stay apart (See Fig.~\ref{fig:our-model}).
We use the learned similarity to study the clustering of generative models in our fingerprint embedding space. 
A particular question we study using this similarity metric is whether
the models in the same family of generative models remain similar in the fingerprint embedding space, or if a different clustering of models emerges.
To answer this question, we use the distance (e.g., Euclidean distance) in the learned latent space as metric in a standard hierarchical clustering algorithm and discover latent families of generative models.
Our experiment in Sec.\ref{sec:exp2} demonstrates there are clear clusters that group VAEs and GANs separately, 
as well as a nested clustering within GANs that partitions the GAN models into  \{StyleGAN, StyleGAN2\}, \{ProGAN, SNGAN\}, \{WGAN-gp, WGAN-lp, MMDGAN, InfoMAxGAN\}, and \{pGAN, EBGAN, DCGAN\}. 

\subsection{What differentiates our work}

Our work is different from existing work on model fingerprints in both conceptual and computational aspects.
Conceptually, our work emphasizes that the extracted fingerprints can serve as a \textit{representation} of the underlying generative models that defines the similarities of the models as learned via contrastive loss (Fig.\ref{fig:our-model}).
This viewpoint is different from existing works \cite{Cozzolino2020NoiseprintAC, Wang2022GANgeneratedFD, Marra2019DoGL, Yu2019AttributingFI} whose main purpose of extracting fingerprints is for a discriminative task such as DeepFake detection or source attribution. 
Although we are interested in such applications, 
we put equal emphasis on the task of using the representations to derive a distance metric between the underlying processes so that we can compare the models quantitatively, which has broader implications than using them only for discriminative tasks. 

Another perspective that distinguishes our work is that it looks at a representation of a \emph{set of images} to extract a model's fingerprint, using a set-based encoder, rather than operating on a single image, as in previous works.
Our experiments  (Sec.~\ref{sec:exp1}, \ref{sec:exp3}) demonstrate that such a set-based method contributes to improved stability of model fingerprints and improved accuracy in source attribution.

From a computational perspective, our method is unsupervised (i.e., does not require ground-truth fingerprint as target labels) and is independent of a fixed downstream, classification task. 
In comparison, \cite{Yu2019AttributingFI} proposed a fingerprint-extraction method that is tied to the downstream task of source attribution.  
Instead, our method leverages the intuition that images from the same model should reflect the features of the same underlying generative process, and thus should be mapped to close representations in the embedding space. 
Our model is trained in an unsupervised setting using contrastive loss \cite{Bromley1993SignatureVU, Chopra2005LearningAS, Cozzolino2020NoiseprintAC} that implements the constraint that representations of (a set of) images generated by the same model should be similar,  while images from different models remain further apart in the embedding space.

We  extend the work on GAN fingerprints to the general class of generative classes. We demonstrate their existence and identifiability, propose a new set-based method to improve the stability and accuracy in source attribution, and use the implicitly learned metric in the fingerprints space as similarity measure of the underlying models to discover latent grouping of models.
\paragraph{Existence and uniqueness of artificial fingerprints} We  extend the existing hypothesis on the existence of GAN fingerprints to a broader class of generative models that supersets the GAN family. 
Specifically, we demonstrate that unique fingerprints exist for VAEs, Flows and score-based models using correlation analysis. 
We then demonstrate that they successfully capture unique and discriminative features of the underlying models by evaluating the fingerprint representations for two downstream tasks, detection of AI-generated images and source attribution task (Sec.~\ref{sec:exp1}). 
\paragraph{Set-based encoding of Fingerprints}  We propose a new set-based encoding of fingerprints and investigate the stability of model fingerprints through a thorough ablation study and comparison to existing methods.
As an ablation study, we investigate how the number of images in an input set (which we define as a \emph{bag} of images in Sec.~\ref{sec:our-method}) affect the stability of model fingerprints. 
Similarly, we study how the number of bags used per model fingerprint affect the stability of the representation.
We evaluate the stability in the degree of correlation and prediction task performance, and 
find that our method achieves stable fingerprints with ~300 images per bag and ~100 bags per model, without compromising its discriminative power. 
In addition, we demonstrate that our set-based encoder contributes to an improved stability of the representation through comparisons to existing methods based on a single-image encoder.

\paragraph{Measure of similarity between generative models} In addition to the discriminative aspects of fingerprints, we investigate the value of the representation as a way to measure similarities between generative models. 
Despite the interest and need for a standard way of comparing/evaluating generative models, there is yet no single metric that can quantify the similarities of generative models, especially in terms of the quality of samples they generate. 
In this work, we propose to use contrastive loss to learn the fingerprint representation so that the distance in the learned latent space implicitly captures the similarity between the models for comparing generative models. 
We perform hierarchical clustering analysis in the learned latent space (i.e., model space) to discover latent groups of generative models based solely on to the samples they generate.

\section{Generalized Hypothesis on Artificial Fingerprints} \label{our-conjecture}

Marra et al \cite{Marra2019DoGL} and Yu et al \cite{Yu2019AttributingFI} 
first proposed a hypothesis on the existence and uniqueness of fingerprints on GANs, 
and provided evidence for their existence and attributability for several well-known GAN models. However, it remains unanswered whether the hypothesis holds on many other types of generative models.  
We extend the scope of this hypothesis to a broader class of generative models that includes VAEs, flows and score-matching models, and propose the following conjecture on the existence of artificial fingerprints in a generalized class of generative models.

\begin{conjecture}
There exists  unique fingerprints of different families of generative models that encompasses VAEs, Flows, GANs and Score-based models.
\end{conjecture}

In Sec.\ref{sec:exp1}, we verify the existence and identifiability of model fingerprints using correlation analysis and evaluation on the task of source attribution. 
In Sec.\ref{sec:exp2}, we demonstrate how the learned fingerprint representations can be used to define a similarity measure between underlying generative models.

\section{Contrastive Sets For Fingerprinting Generative Models} \label{sec:our-method}

\begin{figure}[bt!]
    \centering
    \includegraphics[width=\textwidth]{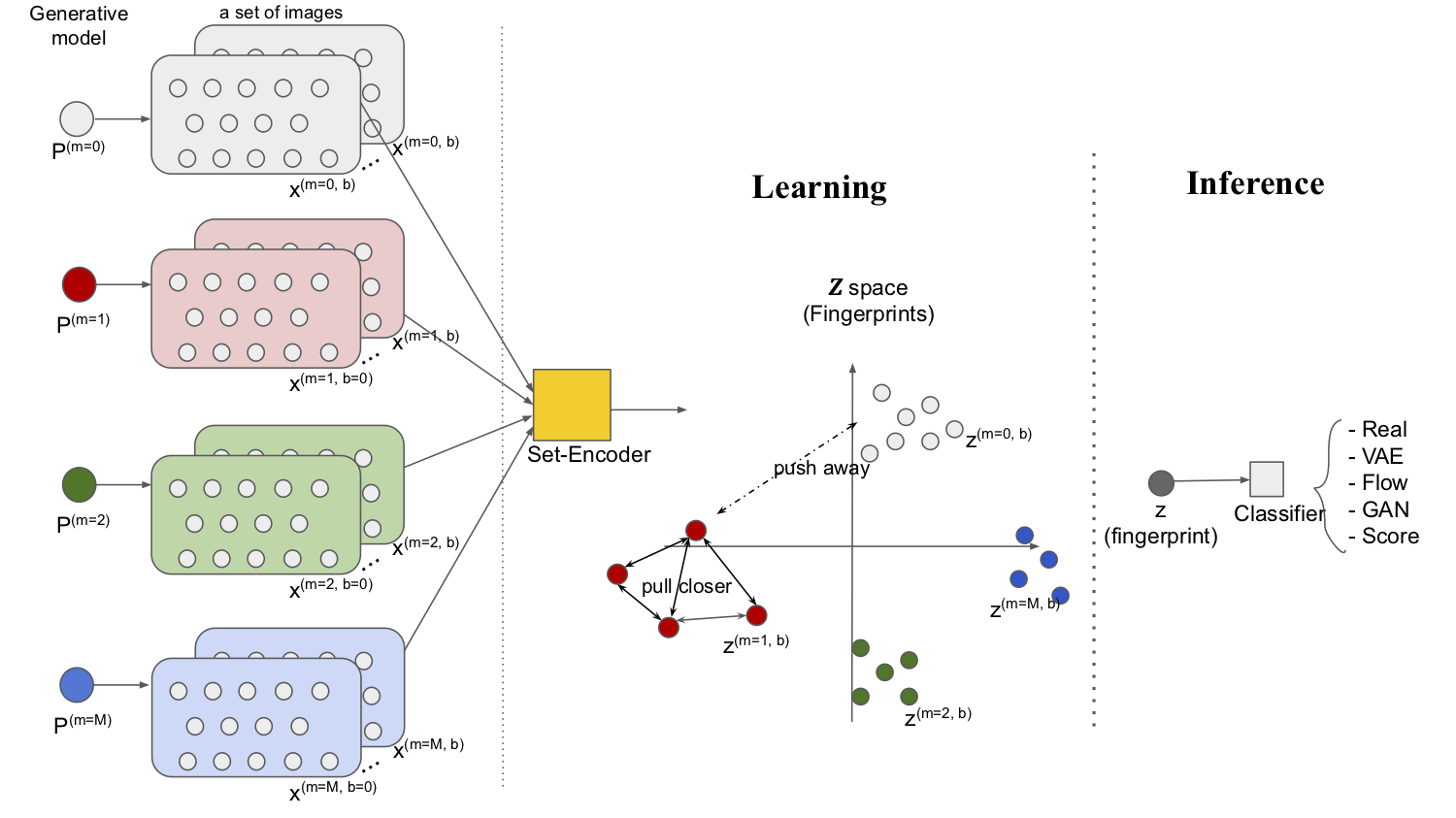}
    \caption{We train a set-encoder with an input of a set of images (a ``bag") with a contrastive loss on the embeddings:
    $z^{(m,b)}$'s from the same generative model $P^{(m)}$ are encouraged to stay close, while $z^{(m,b)}$'s from different generative models are encouraged to stay apart. 
    At inference, we can use the learned fingerprints to solve a downstream task such as the source attribution task in the figure.}
    \label{fig:our-model}
\end{figure}

Motivated by the evidence that generative models leave distinct and consistent fingerprints on the images  they generate, 
we use the fingerprints as representations of the underlying generative models. 
Our main goal with such representations is two-fold.
First, fingerprints capture informative features of the models that uniquely identify the generative models that generated the observed data. 
Secondly, the learned latent space of fingerprints can be used to define a metric that quantifies the similarity between the generative models in regards to the quality of their samples.
Our approach of learning robust fingerprints of generative models consists of two key components: 
a set-encoding network and the contrastive learning framework.

\textbf{Set-encoding network}
A good representation of a model (or equivalently, the sampling distribution of the model) should be learned from a sufficiently large number of images that are sampled from sufficiently diverse parts of the distribution, rather than from a single image.  
To explicitly implement this idea, we propose a set-encoder network that learns model fingerprints from a \textit{set} of images, which we refer to as a \emph{bag of images}. 
More concretely, our set-encoder $F$ receives as an input a bag of images $\mathbf{x^{(m,b)}} = \{ x^{(m,b)}_{1},\dots, x^{(m,b)}_{N}\}$ 
that contains $N$ images generated by the same model $P^{(m)}$,
and outputs a code vector $z$ as the set fingerprint. 

We implement our set-encoder using the DeepSet network with mean pooling as a permutation-invariant operation \cite{Zhang2019DeepSP}.
In other words, our set-encoder $F$ is a composition of an image-level encoder $\phi$ and a permutation-invariation operator $\rho$ such that the following is satisfied:
$F (\mathbf{x}) = (\rho \circ \phi)(\mathbf{x}) = \rho (\sum_{x \in \mathbf{x} } \phi(x)) $, 
where $\mathbf{x}$ is a bag of images. Each image $x_n$ in a bag $\mathbf{x}$ is transformed into some feature $\phi(x_n)$, and the image-level features $\phi(x_n)$ are added up and processed by the $\rho$ network to output a fingerprint for the bag.

\textbf{Contrastive learning}
The second component of our method is contrastive set learning. 
Unlike the standard contrastive loss computed with the features of individual images, 
we apply the contrastive loss on the features of \textit{sets} of images (i.e., the bags).
The contrastive loss encourages the embeddings of bags from the same model to stay close
while those from different models to stay apart.  
We use the squared Euclidean distance to compute the distance between two embeddings.
More formally,  let $\mathbf{I(z,z')} = 1$ if and only if $z$ and $z'$ are embeddings of bags from the same model, and 
let $d(z, z') = ||z - z'||_{2}^{2}$ be the distance between two embeddings of bags, $z$ and $z'$.
We require such distances to be small when $z$ and $z'$ are embeddings of bags from the same model (i.e, $\mathbf{I(z,z')} = 1$, and large otherwise. 
Then, we can derive a suitable probability distribution using a softmax function as,
\begin{equation}
    p_{z}(z') = \frac{e^{-d(z,z')}}{ \sum_{z' \neq z} e^{-d(z,z')} }.
\end{equation}
With this definition, our contrastive loss $L(z, z')$ is defined as follows: 
\begin{equation}
    L(z, z') = - \log \sum_{z':\mathbf{I(z,z')}=1} p_{z}(z'). \label{eqn:our-loss}
\end{equation}

Given a mini-batch of bags from models in the training data, our model is trained to minimize $L$ as defined in \ref{eqn:our-loss}.
Note that, in practice, a mini-batch of size $BS$ provides a lot more pairs of bags that can be used to compute the loss than $BS/2$. This is because any possible combination of the embeddings can be used to compute the contrastive loss as long as we know whether they come from the same model or not. 
Also, note that, we have flexibility in choosing the number of images in a bag
and  the number of bags per model.  
In Sec.~\ref{sec:exp3}, we show that our fingerprint becomes stable and its goodness, as measure by the decorrelation score and the accuracies on the source attribution task, saturates within 300 images per bag and 100 number of bags per model.

\section{Experimental Evaluation} \label{sec:exps}
\subsection{Existence and uniqueness of fingerprints in general}  \label{sec:exp1}
We first verify our generalized hypothesis (Sec.~\ref{our-conjecture}) that generative models beyond GANs leave unique model fingerprints by investigating the fingerprints of VAEs, Flows and score-based model. 
For GANs, we verify if the hypothesis holds true even for more recently developed models such as SN-GAN, StyleGAN and StyleGAN2.

\textbf{Our dataset of generative models}
To verify our generalized hypothesis on the existence and uniqueness of model fingerprints,
we first design a new dataset of generative models, consisting of various trained instances of generative models and their samples, which is better suited for testing our hypothesis. 
The experiments done in \cite{Yu2019AttributingFI} and \cite{Marra2019DoGL} use a mix of GANs trained for different tasks (e.g., progGAN is trained for image synthesis, whereas StarGAN and CycleGAN are trained for image2image transfer) and trained on different datasets (e.g., the (source, target) datasets used for training StarGAN are different from the (source, target) datasets used for CycleGAN). 
Such variations in training datasets and target tasks change the semantics of the model fingerprints (as suggested in \cite{Yu2019AttributingFI}), 
and thus make it hard to compare the quality of fingerprints as feature representations of the models themselves. 
Therefore, in order to study the relationship between the models and their fingerprints on a more equal footing, we conduct this part of experiments by training all models with a fixed dataset (CelebA) using the same preprocessing procedure, and the same target task of image synthesis (rather than a mix of image synthesis and image2image transfer).
Additionally, we make sure the whole training pipeline uses the same data preprocessing steps, in order to exclude their effects on the fingerprints. 
This results in models that generate images that are spatially aligned across all models. 
Such collection of models trained through a consistent data preprocessing pipeline will be useful for future studies on model fingerprints.

See Appendix~\ref{app:our-dataset-generation} for details on the variations of models (model classes, architectures, hyperparameters) and the sampling processes we used to generate our dataset of generative models.

\begin{table}[h!]
  \centering
  \caption{List of models used to generate our fake dataset based on CelebA}
\label{tbl:our-fake-dset}
\begin{tabular}{p{0.1\textwidth}p{0.8\textwidth}}
    \toprule
    Class      & Models \\ 
    \midrule
    VAE        & $\beta$-VAE\cite{Higgins2017betaVAELB}, DFC-VAE\cite{Hou2017DeepFC}, FactorVAE\cite{Kim2018DisentanglingBF}, InfoVAE\cite{Zhao2017InfoVAEIM}, NVAE\cite{Vahdat2020NVAEAD}                \\
    \midrule
    GAN        & plain GAN (pGAN)\cite{Goodfellow2014GenerativeAN}, EBGAN\cite{Zhao2016EnergybasedGA}, DCGAN \cite{Radford2016UnsupervisedRL}, LSGAN\cite{Mao2017LeastSG}, WGAN-gp/lp\cite{Gulrajani2017ImprovedTO}, DRAGAN-gp/lp\cite{kodali2017convergence}, StyleGAN\cite{Karras2019ASG}, StyleGAN2\cite{Karras2020AnalyzingAI}, ProGAN\cite{Karras2018ProgressiveGO}, MMDGAN\cite{Binkowski2018DemystifyingMG}, SNGAN\cite{Miyato2018SpectralNF}, InfoMaxGAN\cite{Lee2021InfoMaxGANIA} \\ 
    \midrule
    Flow       & GLOW\cite{Kingma2018GlowGF}, MaCow\cite{Ma2019MaCowMC}  \\ 
    \midrule
    Score & Noise Conditional Score Network (NCSN) \cite{Song2019GenerativeMB}, Guided Diffusion \cite{Dhariwal2021DiffusionMB}                      \\ 
    \bottomrule

\end{tabular}

\end{table}

\begin{figure}[tb]
\centering
\begin{subfigure}[b]{0.30\textwidth} 
    \centering
    \includegraphics[width=\textwidth]{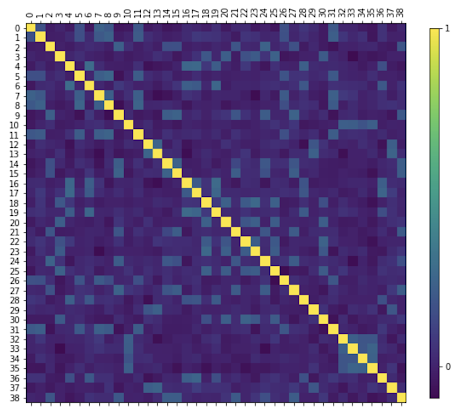}
    \caption[]%
    {{\small Correlation of fingerprints extracted by \cite{Marra2019DoGL} }}    
    \label{fig:exp1-correl-baseline-prnu}
\end{subfigure}
\hfill
\begin{subfigure}[b]{0.30\textwidth}
    \centering
    \includegraphics[width=\textwidth]{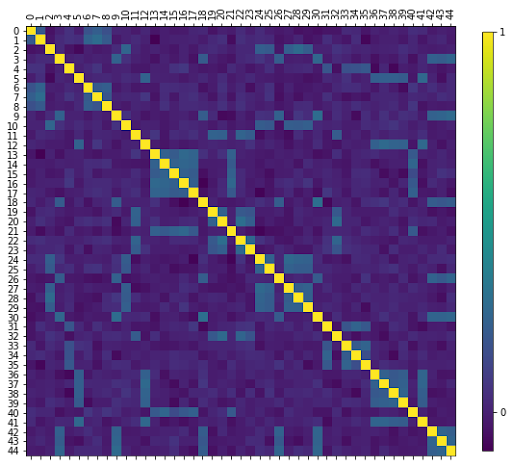} 
    \caption[]%
    {{\small Correlation of fingerprints by our model $n=1$ }}    
    \label{fig:exp1-correl-ours-ae}
\end{subfigure}
\hfill
\begin{subfigure}[b]{0.30\textwidth}
    \centering
    \includegraphics[width=\textwidth]{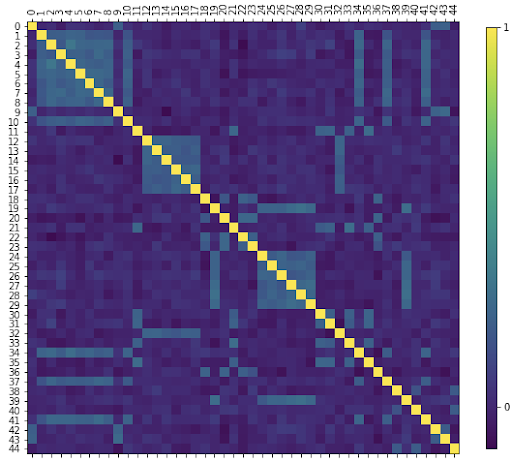}
    \caption[]%
    {{\small Correlation of fingerprints by our model with $n=100$}}    
    \label{fig:exp1-correl-ours-um}
\end{subfigure}
\caption{Correlation analysis of fingerprints from PRNU \cite{Marra2019DoGL} and our set-based methods.}
\label{fig:exp1-correl-results}
\end{figure}

We show evidence of the existence and uniqueness of model fingerprints using correlation analysis, and demonstrates their attributability via evaluation on the downstream task of predicting the source model given samples.

\textbf{Existence and uniqueness of fingerprints for generative models: Correlation analysis}

We measure how close the $r^{{th}}$ model's $i^{{th}}$ image is to the $c^{{th}}$ model's fingerprint using the average correlation index.
Let $\rho_{i,j}^{(n)}$ denote the absolute value of the correlation between $x^{(i,n)}$ (i.e., the $n^{{th}}$ image generated by Model $i$ and the fingerprint of Model $j$):   
\begin{equation}
    \rho_{i,j}^{(n)} = | \mathrm{corr} (\tilde{x}^{(i,n)}, z^{(j)}) |
\end{equation}
We then measure how correlated two model fingerprints are by computing the average of the absolute correlation over the images:
\begin{equation}
    \hat{\rho}_{i,j} =  \frac{1}{N} \sum_{i=1}^{N} | \mathrm{corr} (\tilde{x}^{(i,n)}, z^{(j)}) |
\end{equation}
where the images come from a held-out set of images that are not used for estimating the model fingerprints, and $N$ is the size of this held-out set.
We compute the average correlation between all sets of fingerprints $\{ z_m\}_{m=1}^{M}$, which gives an $(i,j)^{{th}}$ entry of the correlation matrix $S$ (Fig.~\ref{fig:exp1-correl-results}):
\begin{equation}
    \rho_{i,j} =  \frac{1}{N} \sum_{i=1}^{N}  \left| \mathrm{corr} (\tilde{z}^{(i,n)}, z^{(j)}) \right|
\end{equation}
With this definition, we propose a new metric called \emph{correlation score} as a way to measure how distinctive each fingerprint is as a unique representation of its underlying model.
We define the correlation score as the normalized Frobenius norm of the difference between an identity matrix and the average correlation matrix. 
\begin{equation}
    \mathrm{Decorrelation~Score} = \frac{1}{M^2} \| \rho -\mathbf{I} \| 
\end{equation}  
where $M$ is the number of models, and $\mathbf{I}$ is the identity matrix of size $M$. 
Note that each entry $\rho_{i,j}$ is in range [0,1], and the decorrelation score in $[0, 1+\frac{2}{M}]$. 
Also, note that, a higher score indicates the fingerprints are more decorrelated. 

\begin{table}[!ht]  
\caption{Decorrelation scores of 3 baselines
and our models with or without (DSF+/-) set-encoding on GAN dataset \cite{Yu2019AttributingFI} (Left) and our GM dataset (Right)}
\label{tbl:exp1-correl-results}
    \centering
    \begin{tabular}{cc|c} 
    \toprule ~ & Decorrel. Score (GANs) & Decorrel. Score (GMs) \\ 
    \midrule
        PRNU \cite{Marra2019DoGL} & $0.392\pm0.091$ & $0.212\pm0.082$  \\ 
        Yu et al \cite{Yu2019AttributingFI} & $0.549\pm0.079$ & $0.454\pm0.072$  \\ 
        Asnani et al \cite{Asnani2021ReverseEO} &$ \underline{0.645}\pm \underline{0.081}$ & $0.356\pm 0.081$  \\ 
        DSF- (ours) & $0.625\pm0.065$ & $\underline{0.541}\pm\underline{0.078}$ \\ 
        DSF+ (ours) & $\mathbf{0.714\pm0.042}$ & $\mathbf{0.593\pm0.067}$ \\ 
    \bottomrule
    \end{tabular} 
\end{table}

\begin{table}[!ht]  
\caption{Classification accuracy of 3 baselines 
and our models with or without (DSF+/-) set-encoding on GAN dataset \cite{Yu2019AttributingFI} (Left) and our GM dataset (Right)}
\label{tbl:exp1-src-results}
    \centering
    \begin{tabular}{ccc|cc} 
    \toprule ~ &  Acc & AUC (\%) & Acc & AUC (\%) \\
    \midrule
        PRNU \cite{Marra2019DoGL} & $0.785\pm0.135$ & $61.3\pm1.54$ & $0.608\pm0.22$ & $57.4\pm0.089$ \\ 
        Yu et al \cite{Yu2019AttributingFI} & $0.856\pm0.324$  & $88.3\pm2.43$ & $0.712\pm0.072$  & $60.1\pm0.094$ \\ 
        Asnani et al \cite{Asnani2021ReverseEO} & $0.718\pm0.231$ & $76.5\pm1.65$ & $0.708\pm0.453$ & $64.3\pm0.063$ \\ 
        DSF- (ours) & $ \underline{0.903}\pm\underline{0.269}$ & $ \underline{90.5}\pm\underline{1.54} $ & $\underline{0.722}\pm\underline{0.424}$ & $\underline{70.1}\pm\underline{0.059}$\\ 
        DSF+ (ours) & $ \mathbf{0.931\pm0.156} $ & $ \mathbf{93.7\pm1.01}$  & $\mathbf{0.738\pm0.231}$ & $\mathbf{72.9\pm0.043}$ \\ 
    \bottomrule
    \end{tabular} 
\end{table}

\textbf{Attributability of fingerprints}
To evaluate the quality of fingerprints as feature representations of models, we test their accuracy on the source attribution task:
given image samples, we measure how accurately we can identify the source model using the fingerprint representations. High accuracy indicates that the fingerprints encode unique and informative features of the models that allow a reliable discrimination of the models.
Here, we train a downstream classifier with the standard cross-entropy loss, which,  given an input of a fingerprint (i.e.,  an output of the our set-encoding network), it aims to output the correct label of the source model $ y \in  \{Real, G_1, G_2, \dots, G_M \}$.
We implement the classifier with three fully-connected layers. 
We compare our method to 4 baselines, including Inception features \cite{Szegedy2015GoingDW} and three state-of-the-art fingerprint-extractors: PRNU-based method\cite{Marra2019DoGL},  Yu et al \cite{Yu2019AttributingFI} and Asnani et al\cite{Asnani2021ReverseEO}.
We use the classification accuracy to evaluate their source attribution performances.

Table~\ref{tbl:exp1-src-results} shows the classification accuracies of our methods and baselines on the source attribution task. 
We test on two datasets, the dataset of four GANs as in \cite{Yu2019AttributingFI}, and our new dataset of more diverse generative models. 
We report the accuracies and corresponding standard deviations from a 10-fold cross validation. 
Among the methods being compared, our method achieves the highest accuracies with the smallest variances on both datasets. 
Note that the accuracies on the GAN dataset are higher than those on the GM dataset as GAN dataset has fewer labels and thus is an easier classification task.  
Note also that our model with set-encoding (DSF+) achieves a higher performance with a smaller variance than our model without set-encoding (DSF-).

\subsection{Similarity measure for generative models based on fingerprints} \label{sec:exp2} 
\begin{figure}[ht]
    \centering
    \includegraphics[width=\textwidth]{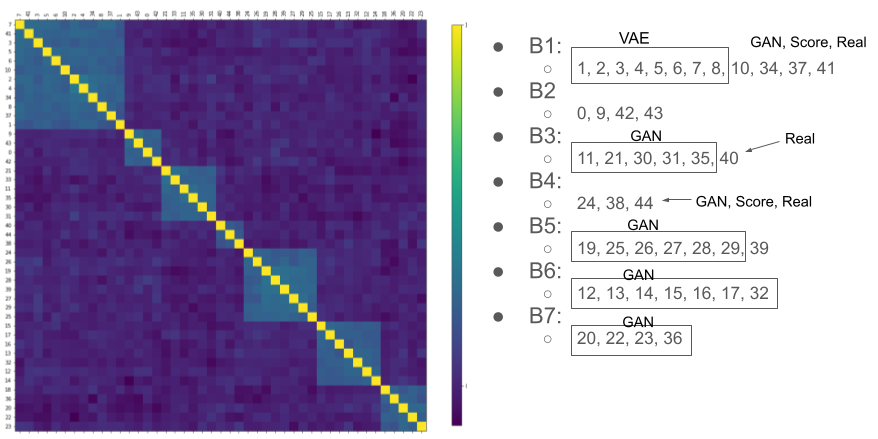}
    \caption{Clusters of models in the learned embedding space. We run a hierarchical clustering on the learned model fingerprints using the correlation score in Sec.\ref{sec:exp1}. Note a clear block-wise pattern, which reflects the clusters of models that are closely correlated with each other in the fingerprint feature space. 
    (Right) Labels of the model class in each block $B$.  }
    \label{fig:exp2-clusters-in-z-space}
\end{figure}

We study the clusters of model families in the latent space to investigate if the similarities of models based on their fingerprints align with the standard categorization of generative models into  \{VAE, GAN, Flow, Score\}. 
In particular, we investigate if the fingerprints of the same model class (e.g., VAE) look similar to each other than to models of other classes (e.g., Real, Flow, GAN, Score).  
Note that when training our model, we impose contrastive loss only on the embeddings of the bags of images ($z$'s), and never over model families. 
Therefore, this experiment studies the structure of latent space based on the similarity between models learned implicitly through our contrastive training. 
To discover the latent families of models, we run a hierarchical clustering 
using the correlation score defined in Sec.~\ref{sec:exp1}, which clusters the models with high correlation scores to the same group.
First, we compute pairwise distances between model fingerprints as the correlation scores (Sec.~\ref{sec:exp1}). 
We then compute the linkage matrix and run a hierarchical clustering using the pairwise distance and linkage matrices.
Fig.~\ref{fig:exp2-clusters-in-z-space} shows the resulting clusters of models in the fingerprint space learned from our model. 
On the right of Fig.~\ref{fig:exp2-clusters-in-z-space}, we label the classes of generative models that belong to the same clusters discovered via the hierarchical clustering. 
Note that  there are clear clusters that group VAEs and GANs separately, 
as well as a nestedclustering within GANs that partitions the GAN models into {StyleGAN, StyleGAN2}, {ProGAN, SNGAN}, {WGAN-gp, WGAN-lp, MMDGAN, InfoMAxGAN}, and {pGAN, EBGAN, DCGAN}.
These clusters in the embedding space can be interpreted as latent families of models that are discovered based on the fingerprint representations.

\subsection{Stability of model fingerprints} \label{sec:exp3}
\begin{figure}[t]
\centering
\begin{subfigure}[b]{0.47\textwidth} 
    \centering
    \includegraphics[width=\textwidth]{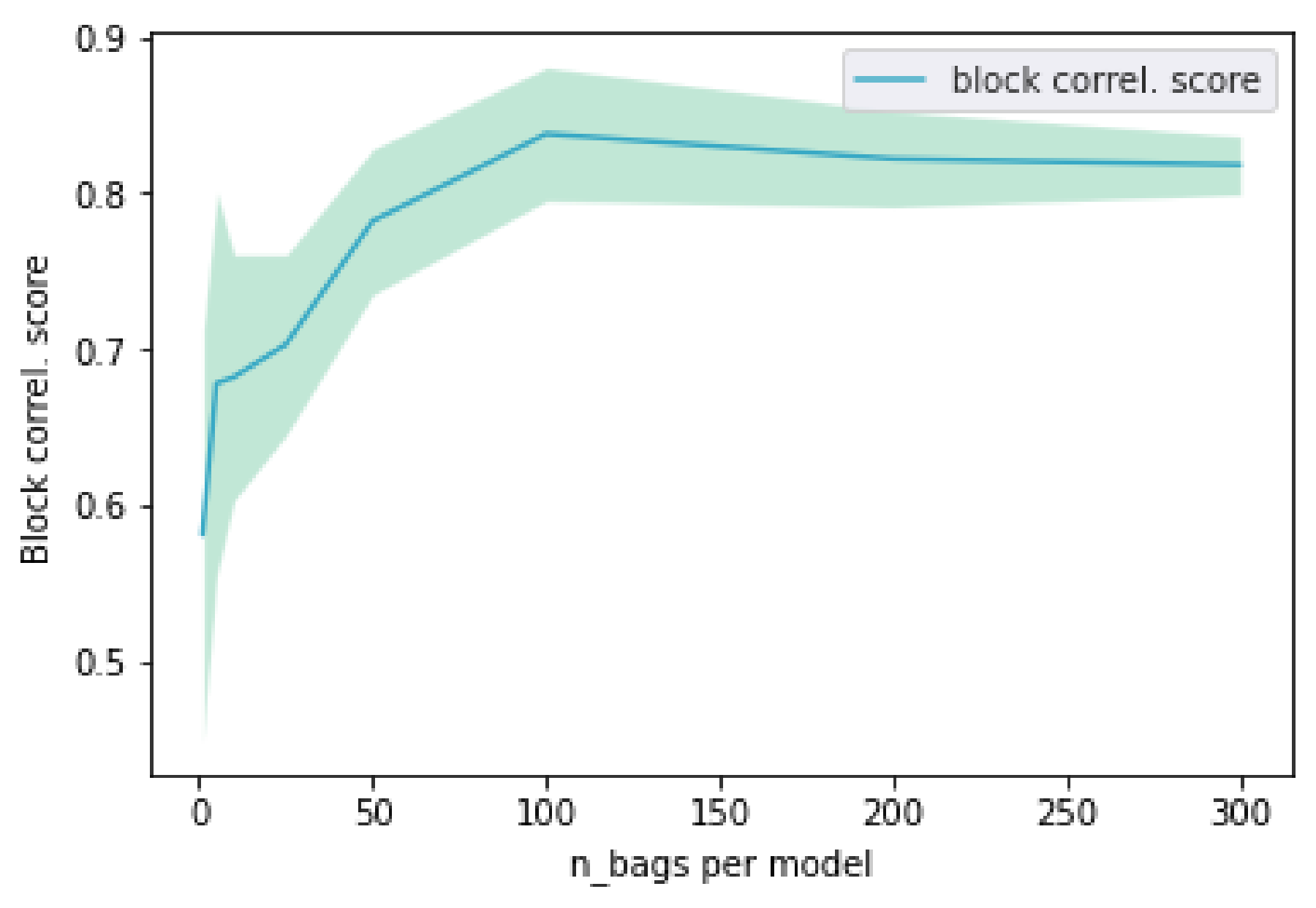}
    \caption[]%
    {{\small Decorrelation score vs. number of bags per model}}
    \label{fig:exp3-block-correl}
\end{subfigure}
\hfill
\begin{subfigure}[b]{0.52\textwidth}
    \centering
    \includegraphics[width=\textwidth]{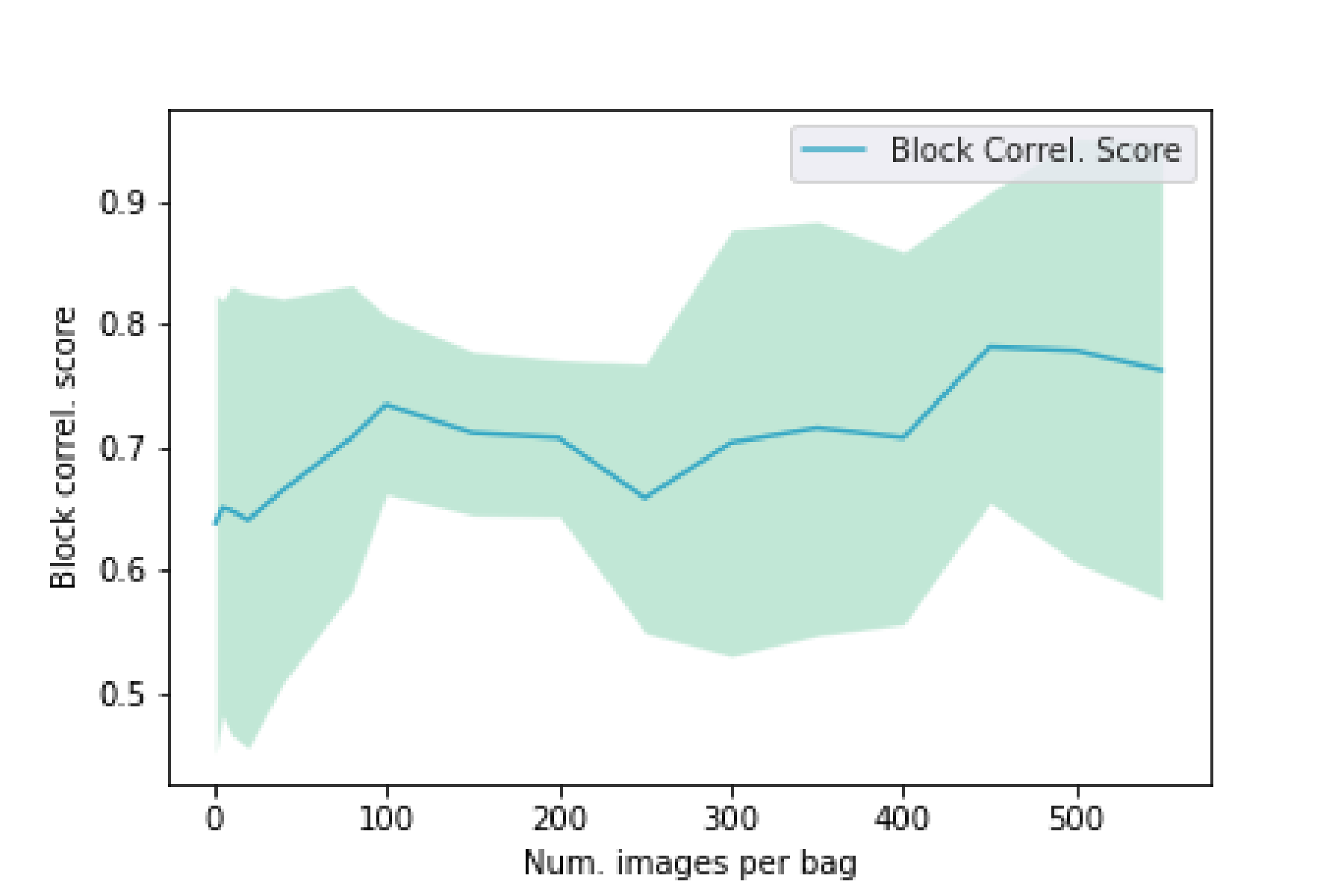}
    \caption[]%
    {{\small Decorrelation score vs. number of images per bag}} 
    \label{fig:exp3-block-score-vs-n-imgs}
\end{subfigure}
\caption{(a) Decorrelation score vs. number of bags per model: as the number of bags 
increase, the decorrelation score increases. Note the score plateaus after ~100 bags per models, which indicates our representation becomes stable in regards to the amount of decorrelation across model instances with a practical size of bags.
(b) Decorrelation score vs. number of images per bag.
}
\label{fig:exp3-stability-plots}
\end{figure}

We study the effects of two key parameters in our method (the number of images per bag, and the number of bags per model) on the stability of the fingerprints. 
We measure the stability with respect to the change in correlation score and the classification accuracies. 
We first investigate how many images in a bag ($n$ in Fig.~\ref{fig:our-model}) are needed to get a stable fingerprint of a model. 
Fig.~\ref{fig:exp3-block-correl} demonstrates that the decorrelation score increases as the number of images per bag increases from 1 to 300.
The standard deviations, marked as the shaded area in Fig.~\ref{fig:exp3-block-correl}, decrease as the bag size increases, which demonstrates how our set-encoder contributes to learning more robust fingerprints by lowering the variance. 
The score plateaus at around $n=100$, while the standard deviation continues to decrease from $n=100, \dots, 300$.

\section{Related Work} 
\subsection{Generative Models and Source attribution}
A variety of recent generative models \cite{Vahdat2020NVAEAD, Kingma2018GlowGF, Song2019GenerativeMB} have successfully achieved high photorealism in image synthesis that are comparable to that of GANs . 
We choose the following four classes of generative models as representative candidates of the current state of the art and study their fingerprints: 
(1) VAEs \cite{Kingma2014AutoEncodingVB, Higgins2017betaVAELB, Vahdat2020NVAEAD, Zhao2019InfoVAEBL}, (2) Flows \cite{Kobyzev2019NormalizingFI, Kingma2018GlowGF}, 
(3) implicit models/GANs \cite{Goodfellow2014GenerativeAN, Radford2016UnsupervisedRL, Li2017MMDGT, Bellemare2017TheCD, Arjovsky2017WassersteinG}, and 
(4) score-based models such as diffusion probabilistic models \cite{SohlDickstein2015DeepUL, Song2019GenerativeMB, Ho2020DenoisingDP, Dhariwal2021DiffusionMB}. 

\textbf{Device fingerprints} 
In traditional visual forensics, it has been studied that different cameras leave unique artifacts on the images, which can be used to detect tampered media \cite{Fridrich2009DigitalIF}. 
Most state of the art methods \cite{Luks2006DigitalCI} have relied on high-frequency residuals to detect the manipulations, yet more recently, CNN-based methods \cite{Bayar2016ADL, Zhou2018LearningRF, Cozzolino2017RecastingRL} have improved the detection performance by learning a more complex, and often imperceptible, features from the manipulated visual media.

\textbf{Artificial fingerprints}  
Prior work takes insights from the idea of the traditional steganalysis and device fingerprints and applies to the problem of identifying GAN-generated images:
\cite{Marra2019DoGL} applies the PRNU-based method from the device fingerprint to study the fingerprints of GANs, and provide evidence that the fingerprints can be used for source identification.
\cite{Yu2019AttributingFI} proposes a CNN-based method to extract fingerprints at the image- and model-level. The authors demonstrate the fingerprints of GANs are unique and useful for identifying the source model. 
Their work however focuses on GANs, and does not address other classes of generative model included in our analysis. 
\cite{Asnani2021ReverseEO} proposes a new problem of ``model parsing'', which estimates various design parameters involved in training a neural network such as the model architecture of neural networks, hyperparameters and loss functions. 
They propose to learn two separate networks, one for extracting fingerprints from images and another for predicting the design parameters, and show the fingerprint extraction pipeline improves the accuracy on the  model parsing task. 

Another group of work focuses on the frequency domain of the images rather than the pixel domain to look for the traces of generative models. \cite{Wang2022GANgeneratedFD},   \cite{Durall2020WatchYU} and \cite{Frank2020LeveragingFA} demonstrate that upsampling operations in CNN-based generative models leave artifacts that become evident in the frequency domain. 
However, subsequent work  has developed GANs that dodge such detectors by better matching to the spectral distribution of the true data, calling for more generalizable and robust fingerprint representations. 
Our approach does not presume the presence of fingerprint to a fixed domain (pixel/frequency) and lets the unsupervised contrastive loss guide the learning of fingerprints (Sec.\ref{sec:our-method}). 

Our work extends the existing studies on GAN fingerprints to a broader class of generative models and demonstrate their existence and attributability in general.  
Concretely, we propose a Generalized Fingerprint Hypothesis \ref{our-conjecture} that not only GANs, but also other classes of generative models manifest unique fingerprints on their samples.
We study the artificial fingerprints of the three prominent families of deep generative models: (1) likelihood-based models that includes VAEs \cite{Kingma2014AutoEncodingVB, Higgins2017betaVAELB, Vahdat2020NVAEAD, Zhao2019InfoVAEBL} and Flows \cite{Kobyzev2019NormalizingFI, Kingma2018GlowGF}, 
(2) implicit models/GANs \cite{Goodfellow2014GenerativeAN, Radford2016UnsupervisedRL, Li2017MMDGT, Bellemare2017TheCD, Arjovsky2017WassersteinG}, and 
(3) score-based models such as Diffusion probabilistic models \cite{SohlDickstein2015DeepUL, Song2019GenerativeMB, Ho2020DenoisingDP}.

\subsection{Measure of similarity between generative models} 
\textbf{Methods based on semantic features}
Current  methods of comparing generative models based on their samples rely on the semantic features of the samples. 
Fréchet Inception Distance (FID) \cite{Heusel2017GANsTB} approximates the Wasserstein metric between distributions using the features of images extracted from a pre-trained network such as the Inception v3 \cite{Szegedy2015GoingDW}. 
FID makes an assumption that the underlying distributions are unimodal Gaussians, and uses the estimated mean and covariance matrices of the semantic features. 
Despite its wide use in benchmarking generative models, FID is prone to inaccurate comparisons due to its biased nature with large variance and its Gaussianity assumptions.
 \cite{Chong2020EffectivelyUF, Xu2018AnES, Theis2016ANO, Borji2019ProsAC}.
An alternative metric that is shown to be unbiased with smaller variance is the Kernel-Inception Distance (KID) \cite{Binkowski2018DemystifyingMG}. 
KID computes  a polynomial kernel $k(x,y)  =  (\frac{1}{d} x^{T} y+ 1)^3$ and measures the associated Kernel Maximum Mean Discrepancy (kernel MMD) between samples from the two  distributions under comparison. 
It is motivated from the Kernel-based two-sample tests \cite{Gretton2012AKT} and thus is suitable for testing which of the two models is closer to the true data distribution  \cite{Sutherland2017GenerativeMA, Bounliphone2016ATO}. 
Since KID replies on a kernelized feature representation of the samples in infinitely many dimensional space, it is hard to interpret the features unlike our fingerprints in finite dimensions. 

It is important to note that both FID and KID are used as a tool to evaluate the \textit{fit} of a generative model to the fixed reference data distribution (e.g., training dataset), 
rather than as a measure of \textit{similarity} for comparing \textit{multiple} generative models. 
    
\textbf{Distance based on fingerprint representation}
While these methods aim to approximate the distance between models using the \textit{semantic} representation of their samples, such semantic representation may not be effective in distinguishing high-capacity generative models that achieve comparable photorealism in image synthesis. 
For this reason, our work takes a different approach and uses the non-semantic yet discriminative artifacts left on the samples as the basis of similarity measure between the underlying models.

\clearpage
\bibliography{00-mainbib}
\bibliographystyle{ref}

\clearpage
\appendix

\section{Appendix} 
\subsection{Discrepancies in data-preprocessing in existing training scripts} 
\textbf{Effect of using different data-preprocessing when training generative models}
Upon inspection of the source codes of the generative models of our interest (See Table~\ref{tbl:our-fake-dset}, 
we noticed the discrepancies in their data pre-processing pipelines. 
For example, Ho\cite{Ho2021GitrepoGANs} trains various GAN variants \cite{Goodfellow2014GenerativeAN, Radford2016UnsupervisedRL, kodali2017convergence, Arjovsky2017WassersteinG, Mao2017LeastSG} on CelebA using the center-crop at 128,
while \cite{Song2021GitreproScores} trains their score-based models \cite{Song2019GenerativeMB, Song2021ScoreBasedGM} on CelebA using a center-crop at 140  (followed by a resizing to 64 by 64).

\begin{figure}[H]
\centering
\begin{subfigure}[b]{0.475\textwidth} 
    \centering
    \includegraphics[width=\textwidth]{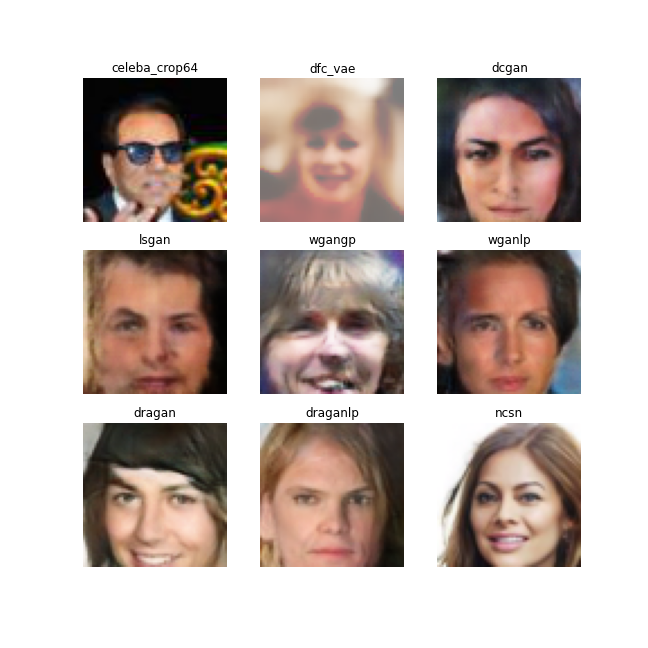}
\end{subfigure}
\hfill
\begin{subfigure}[b]{0.475\textwidth}
    \centering
    \includegraphics[width=\textwidth]{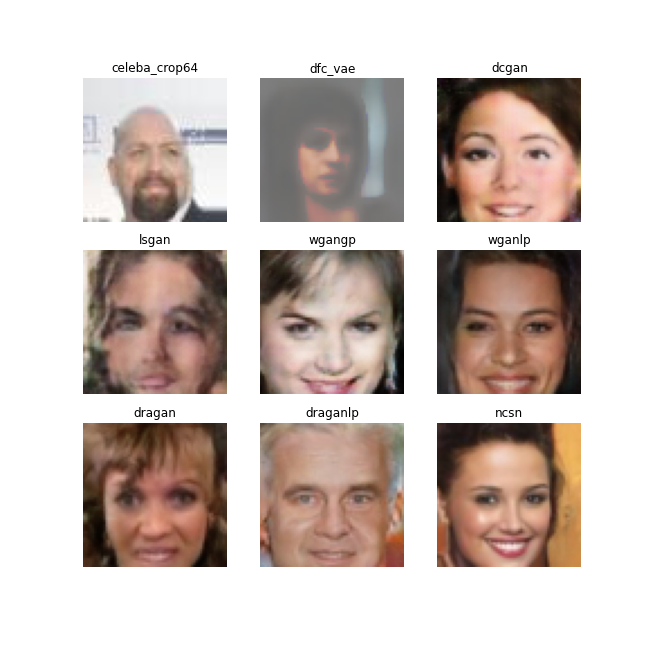}
\end{subfigure}
\vskip\baselineskip
\begin{subfigure}[b]{0.475\textwidth}
    \centering
    \includegraphics[width=\textwidth]{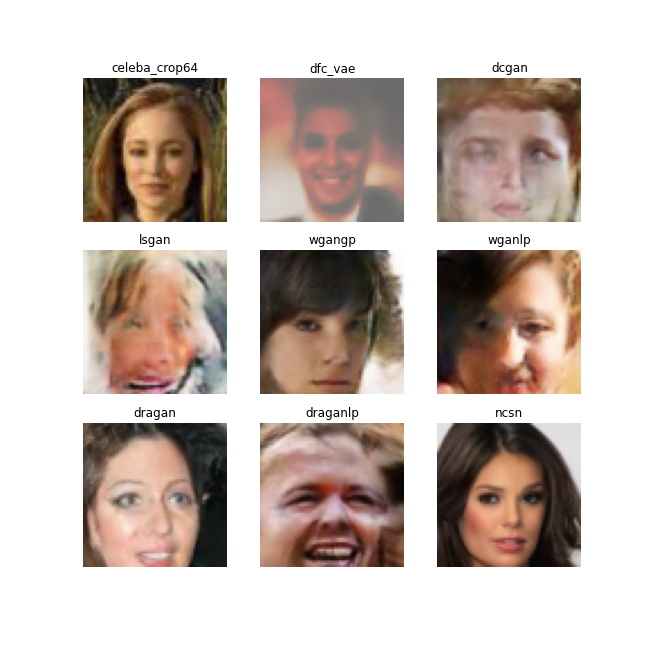}
\end{subfigure}
\hfill
\begin{subfigure}[b]{0.475\textwidth}
    \centering
    \includegraphics[width=\textwidth]{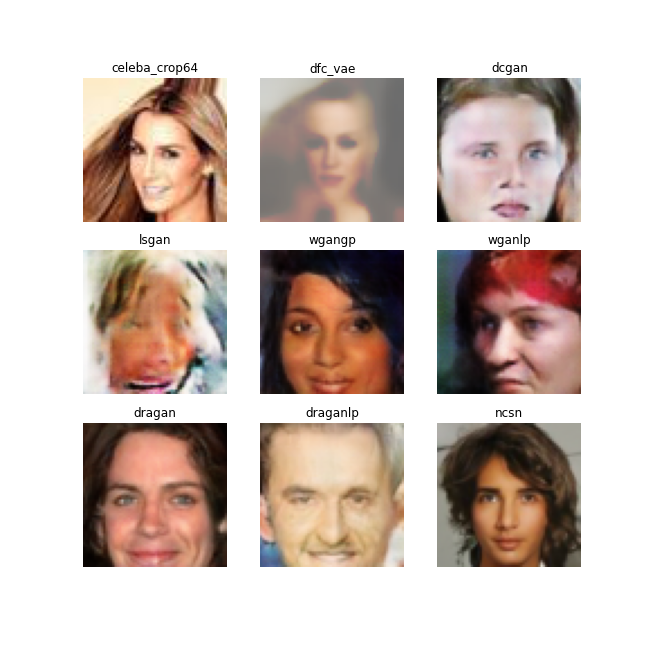}
\end{subfigure}
\caption{Data-preprocessing needs to be consistent to train generative models in our study because our goal is to study the relationship of models and their fingerprints.}
\label{app:images-diff-data-preproc}

\end{figure}

\subsection{Generation of our fake dataset} \label{app:our-dataset-generation}
We create a new dataset of generative models that includes various state-of-the-art models from four representative classes of models, VAEs, Flows, GANs, and score-based models.  See Table ~\ref{tbl:our-fake-dset} for the full list of models.

This dataset contains four types of VAEs, 14 types of GANs, two flow-based models and three types of score-based models which are all trained on the same training data (CelebA \cite{liu2015faceattributes}) with the same data-processing steps. 
We use this dataset of models as the basis of our analysis on artificial fingerprints. 
This consistency in our training protocol allows for the comparison of fingerprints to be attributed to the differences in the models, rather than a combination of models, training data and downstream tasks as in previous works \cite{Yu2019AttributingFI, Marra2019DoGL, Asnani2021ReverseEO}. 


\paragraph{Samples from models in our dataset} We present samples from a subset of trained generative models in our dataset.

\begin{figure}[ht]
\centering
\begin{subfigure}[b]{0.45\textwidth} 
    \centering
    \includegraphics[width=\textwidth]{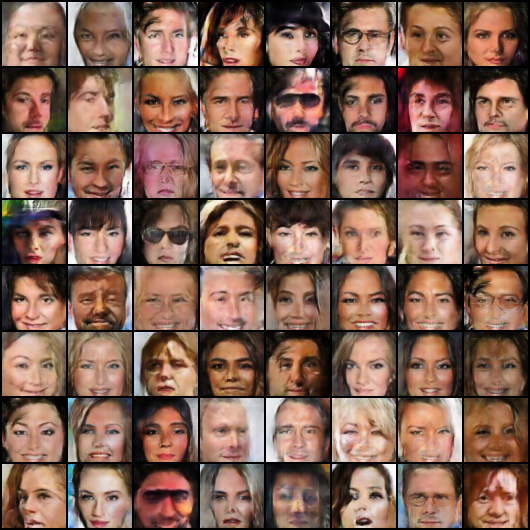}
    \caption[]%
    {{\small DCGAN}}    
\end{subfigure}
\hfill
\begin{subfigure}[b]{0.45\textwidth}
    \centering
    \includegraphics[width=\textwidth]{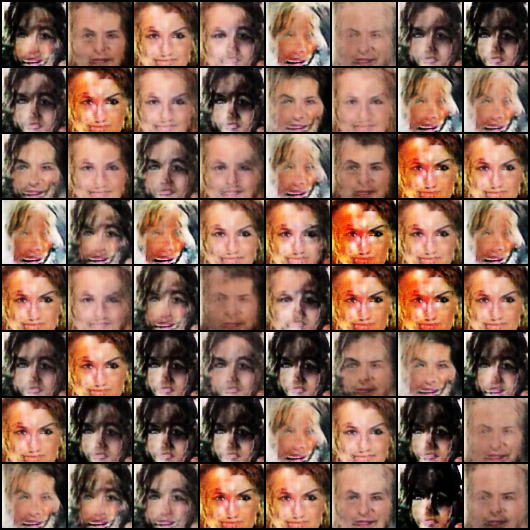} 
    \caption[]%
    {{\small LSGAN}}    
\end{subfigure}
\vskip\baselineskip
\begin{subfigure}[b]{0.45\textwidth}
    \centering
    \includegraphics[width=\textwidth]{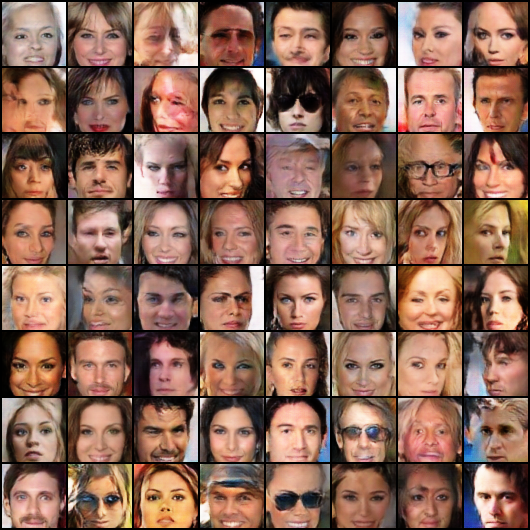} 
    \caption[]%
    {{\small WGAN-gp: gradient penalty, line}} 
\end{subfigure}
\hfill
\begin{subfigure}[b]{0.45\textwidth}
    \centering
    \includegraphics[width=\textwidth]{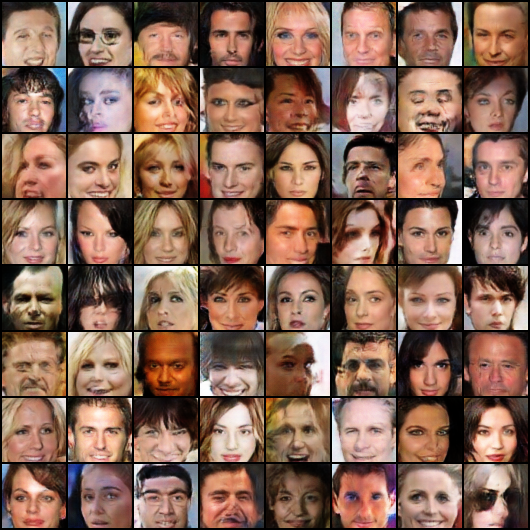}
    \caption[]%
    {{\small WGAN-lp: gradient penalty, line}} 
\end{subfigure}
\vskip\baselineskip
\begin{subfigure}[b]{0.45\textwidth}
    \centering
    \includegraphics[width=\textwidth]{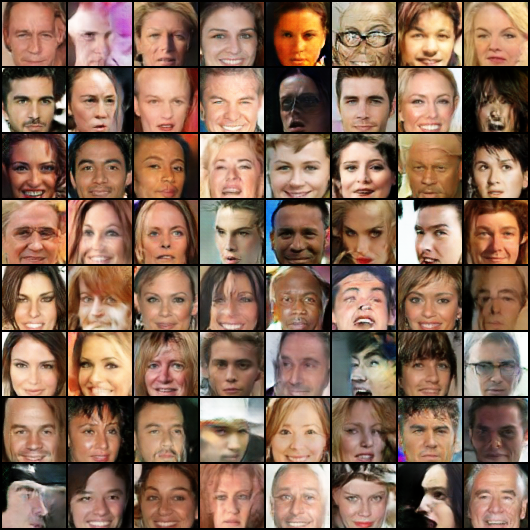}
    \caption[]%
    {{\small DRAGAN, gradient penalty, line}}
\end{subfigure}
\hfill
\begin{subfigure}[b]{0.45\textwidth}
    \centering
    \includegraphics[width=\textwidth]{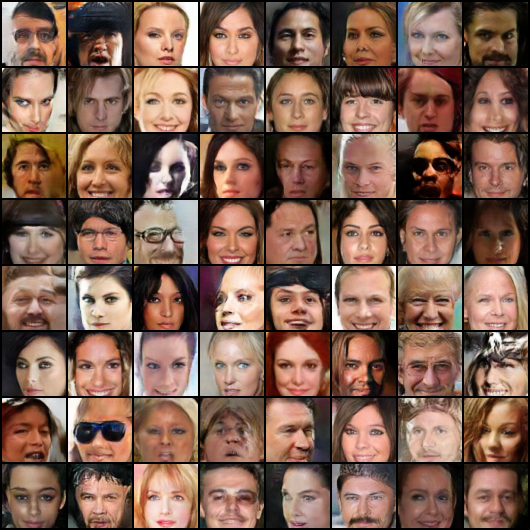}
    \caption[]%
    {{\small DRAGAN, lp, line}}    
\end{subfigure}

\caption{Our fake dataset generated from various GAN models: DCGAN, LSGAN, WGAN-gp, WGAN-lp, DRAGAN-gp, DRAGAN-lp}
\label{fig:our-fakes-gan}
\end{figure}

\subsection{Ablation}
We present the results from our ablation study on the effect of the number of bags per model on the correlation score and the classification accuracy. 

\begin{table}[h!]
    \centering
    \begin{tabular}{cccccccccc}
    \hline
        n\_bags per model & Correl. Score & AUC (\%)  \\ \hline \hline
        1 & $0.581\pm 0.032$ & $55.3 \pm 1.334$  \\ \hline
        5 & $0.678 \pm  0.023$ & $64.4 \pm  1.251$  \\ \hline
        10 & $0.682 \pm  0.019$ & $63.1 \pm  1.023 $ \\ \hline
        25 & $0.703 \pm  0.018$ & $64.7 \pm  1.022 $ \\ \hline
        50 & $0.782 \pm  0.017$ & $72.4 \pm  1.013$  \\ \hline
        100 & \textbf{$0.838 \pm  0.013$} & \textbf{$76.9 \pm  0.873$}  \\ \hline
        200 & $0.822 \pm  0.011 $ & $77.3 \pm  0.822 $ \\ \hline
        300 & $0.818 \pm  0.009 $ & $78.9 \pm  0.723 $ \\ \hline
    \end{tabular}
\end{table}

\end{document}